\documentclass{esannV2}
\usepackage{graphicx}
\usepackage[utf8]{inputenc}
\usepackage{amssymb,amsmath,array}
\usepackage[table]{xcolor}
\usepackage{tikz}
\usepackage{booktabs}
\usepackage{tcolorbox}
\usepackage{caption}
\usepackage{float}
\usepackage{titlesec}
\usepackage{enumitem}
\usepackage{hyperref}
\usepackage{fancyhdr}

\fancypagestyle{esannproceed}{%
  \fancyhf{}%
  \fancyhead[L]{\hspace*{-1.4cm}\parbox{15cm}{\footnotesize ESANN 2026 proceedings, European Symposium on Artificial Neural Networks, Computational Intelligence and \\ Machine Learning. Bruges (Belgium) and online event, 22-24 April 2026, i6doc.com publ., ISBN 9782875870964. \\ Available from http://www.i6doc.com/en/.}}%
}
\begin{document}
\title{High Performance, Low Reliability: Uncertainty Benchmarking for Tabular Foundation Models}

\author{José Lucas De Melo Costa, Fabrice Popineau, Arpad Rimmel, Bich-liên Doan \thanks{This work used HPC resources from the Mésocentre of CentraleSupélec and ENS Paris-Saclay, supported by CNRS and Région Île-de-France, and also access to IDRIS under the GENCI allocation AD011011828R5. It was further supported by the Chair “Artificial intelligence applied to credit card fraud detection and automated trading,” led by CentraleSupélec and sponsored by LUSIS.}
\vspace{.3cm}\\
CentraleSupélec, Université Paris-Saclay \\
Gif-sur-Yvette - France
}

\maketitle
\thispagestyle{esannproceed}

\begin{abstract}
Recent Tabular Foundation Models (TFMs) have demonstrated state-of-the-art predictive performance, often surpassing Gradient-Boosted Decision Trees (GBDTs). However, the trustworthiness of these models, particularly their uncertainty quantification, has been largely overlooked. We investigate this gap through an extensive study comparing TFMs, GBDTs, and classical baselines on the 112 datasets of the TALENT benchmark. Our results reveal a performance–uncertainty trade-off: although TFMs achieve the highest predictive performance (AUC), they exhibit lower conditional coverage under conformal prediction (SSCS) compared to GBDTs. Complementary experiments on synthetic datasets further characterize the regimes in which this effect intensifies. We conclude that while TFMs advance predictive frontiers, achieving well-calibrated uncertainty remains a major open challenge for their reliable adoption. Code is available at: \href{https://github.com/jose-melo/high-performance-low-reliability}{https://github.com/jose-melo/high-performance-low-reliability}
\end{abstract}

\section{Introduction}

Tabular data remain commonly found across industrial and scientific domains, where reliable predictive models are crucial for decision making in areas such as finance \cite{grover_fraud_2023} and healthcare \cite{di_martino_explainable_2023}.
Historically, deep learning methods have struggled to match the performance of well-established Gradient Boosted Decision Trees (GBDTs) when applied to tabular datasets \cite{grinsztajn_why_2022}.

Recently, Tabular Foundation Models (TFMs) have emerged as a promising paradigm, leveraging large-scale synthetic pretraining combined with carefully designed inductive biases \cite{hollmann_tabpfn_2022,hollmann_accurate_2025,qu_tabicl_2025}. Models like TabPFN and TabICL have demonstrated the potential to not only narrow the gap but in some cases surpass classical methods, delivering superior performance on small and medium-scale tabular datasets in terms of accuracy and speed. While recent benchmarks and open leaderboards like \emph{TabArena} \cite{erickson_tabarena_2025} assess the performance of TFMs, they provide little insight into model uncertainty. However, measuring uncertainty is as crucial as raw predictive performance, given that tabular data are often embedded in safety-critical applications.

This lack of systematic analysis leaves open fundamental questions regarding the reliability and deployment of TFMs in critical scenarios. In this work, we address this gap by proposing a comprehensive benchmark of TFMs that evaluates not only predictive performance, but also uncertainty quantification, by the means of conformal prediction. Perpendicular to predictive performance, we introduce a new axis of investigation, providing new insights into the strengths and weakness of Tabular Foundation Models.

Specifically, we benchmark four Tabular Foundation Models against classical and deep learning-based alternatives across the TALENT benchmark \cite{liu_talent_2024} on 112 small and medium tabular datasets, complemented by synthetic datasets designed to isolate the conditions under which uncertainty amplifies. Our empirical results reveal that, although TFMs achieve higher performance, they exhibit greater predictive uncertainty, notably under high-noise and low-separability conditions. \textbf{Main contributions:}
\begin{enumerate}
  \item  We propose an evaluation protocol combining AUC, calibration, and conformal uncertainty metrics;
  \item We conduct a large-scale benchmark comparing TFMs with tree-based, classical and deep learning baselines on tabular data;
  \item We provide empirical evidence that TFMs systematically yield lower size-stratified coverage scores than traditional models, further characterizing exactly when this validity gap widens
\end{enumerate}

\section{Related Work}

\textbf{Tabular Foundation Models} (TFMs) leverage transformer architectures and large-scale synthetic pretraining to approximate Bayesian inference in a single forward pass, achieving state-of-the-art accuracy and efficiency on small and medium tabular datasets \cite{hollmann_tabpfn_2022,hollmann_accurate_2025,qu_tabicl_2025}. Despite their strong performance, prior work has focused almost exclusively on accuracy, offering limited insight into model reliability or calibration.

\textbf{Uncertainty quantification} (UQ) has become a central topic for evaluating the trustworthiness of modern foundation models. Inspired by recent work applying conformal prediction to benchmark uncertainty in large language models \cite{ye_benchmarking_2024}, we adopt the same model-agnostic framework to evaluate TFMs. While conformal methods have proven effective for diagnosing overconfidence in LLMs, their application to tabular foundation models remains unexplored. Our work fills this gap by providing the first systematic assessment of TFM uncertainty and its trade-off with predictive performance.

\section{Methodology}

\begin{figure}
    \centering
    \includegraphics[width=0.9\linewidth]{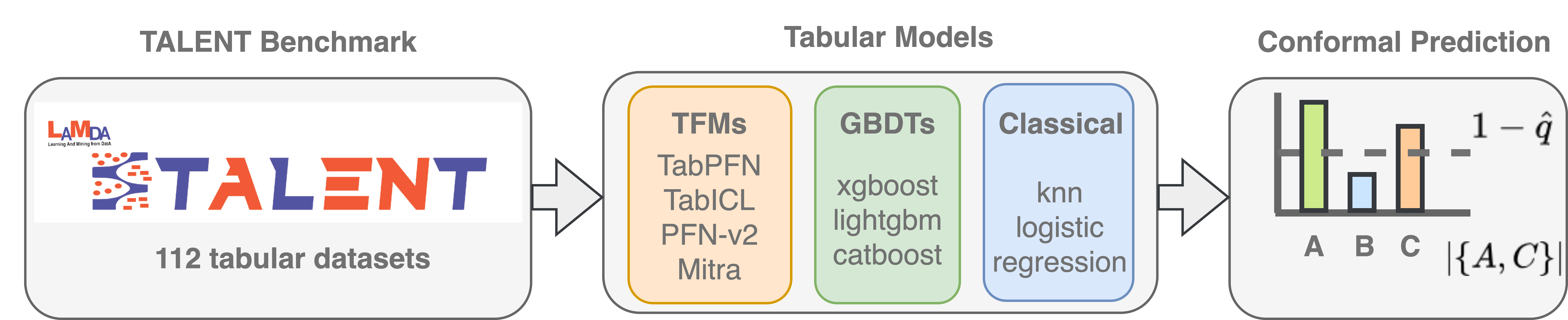}
     \caption{Overview of the proposed methodology. Models from different families (TFMs, GBDTs, and classical baselines) are trained on the TALENT benchmark and evaluated through conformal prediction to derive calibrated prediction sets.}
    \label{fig:overview}
\end{figure}

We evaluated a diverse set of tabular models under a conformal prediction framework, following the protocol illustrated in Figure~\ref{fig:overview}. This setup is inspired by the benchmarking methodology introduced in \cite{ye_benchmarking_2024} for Large Language Models.

\subsection{Conformal Prediction Framework}

To rigorously evaluate the uncertainty of Tabular Foundation Models (TFMs), we adopt the framework of \emph{conformal prediction} (CP). Conformal prediction provides distribution-free guarantees on predictive uncertainty by transforming point predictions into \emph{prediction sets}. These sets are designed to contain the true label with a user-specified probability $\alpha$, irrespective of the underlying data distribution \cite{angelopoulos_gentle_2022}.

Let $\hat{p}(y \mid x)$ denote the class probabilities estimated by a model (e.g., a TFM). Conformal prediction relies on a calibration set $I_{\text{cal}}$ disjoint from the training data. For each $(x,y) \in I_{\text{cal}}$, a \emph{nonconformity score} $s(x,y)$ is computed, reflecting how unusual the label $y$ appears given the prediction $\hat{p}(\cdot \mid x)$. Given the collection of nonconformity scores on $I_{\text{cal}}$, we compute the empirical $(1-\alpha)(1 + 1/|I_{\text{cal}}|)$-quantile $q_\alpha$. For a new instance $x$, the conformal prediction set is then defined as $\hat{C}_\alpha(x) = \{ y \in \mathcal{Y} \,:\, s(x,y) \leq q_\alpha \}$.
By construction, this procedure guarantees that $\Pr\{ y \in \hat{C}_\alpha(x) \} \geq 1 - \alpha$, ensuring valid marginal coverage without distributional assumptions. We use the \emph{Least Ambiguous Class} (LAC) score, $s(x,y)=1-\hat{p}(y \mid x)$ at target coverage $1-\alpha=0.90$.

\subsection{Metrics}

Once prediction sets are generated, their quality is assessed through metrics that jointly capture reliability and informativeness. We first consider the \textbf{Coverage Rate (CR)}, defined as the proportion of test samples for which the true label lies inside the prediction set: $\text{CR} = \frac{1}{n}\sum_{i=1}^n \mathbf{1}\{ y_i \in \hat{C}_\alpha(x_i)\}$.
A well-calibrated model should achieve coverage close to the target level $1-\alpha$.

To quantify how informative these sets are, we use the \textbf{average set size (SS)} $\text{SS} = \frac{1}{n}\sum_{i=1}^n |\hat{C}_\alpha(x_i)|$, which measures the typical cardinality of prediction sets. Smaller sets correspond to more confident predictions, whereas larger sets indicate greater uncertainty. However, informativeness alone is insufficient: a model that produces small but unreliable sets is untrustworthy. To study reliability \emph{conditionally} on uncertainty, we rely on \textbf{Size-Stratified Coverage (SSC)}, which measures empirical coverage across groups of samples with similar set sizes: $\mathrm{SSC}(k) = \frac{1}{|\mathcal{G}_k|}
\sum_{i \in \mathcal{G}_k} \mathbf{1}\{ y_i \in \hat{C}_\alpha(x_i) \}$, where $\mathcal{G}_k = \{\, i : |\hat{C}_\alpha(x_i)| = k \,\}$. The overall score is $\mathrm{SSCS} = \min_{k} \mathrm{SSC}(k)$.

\begin{tcolorbox}[colback=gray!10, colframe=gray!50]
    \textbf{Why conditional coverage matters? } Consider two groups:
    \vspace{-0.5em}
    \begin{itemize}
        \item \textbf{Group A:} Small prediction sets ($|\hat{C}_\alpha(x_i)| = 1$) cover the true label only 60\% of the time.
        \vspace{-0.5em}
        \item \textbf{Group B:} Large prediction sets ($|\hat{C}_\alpha(x_i)| = 5$) cover it 100\% of the time.
    \end{itemize}
    \vspace{-0.5em}
    The overall coverage may still average to the target (e.g., 90\%), yet the model is poorly calibrated: \textbf{confident predictions (small sets) are unreliable}, while uncertain ones are overly conservative.
\end{tcolorbox}

\subsection{Implementation details}

Four families of models were considered: (i) Tabular Foundation Models (TFMs): TabPFN \cite{hollmann_tabpfn_2022}, TabICL \cite{qu_tabicl_2025}, PFN-v2\cite{hollmann_accurate_2025}, and Mitra\cite{zhang_mitra_2025}; (ii) GBDTs: xgboost, lightgbm, and catboost; (iii) classical baselines: $k$-nearest neighbors and logistic regression and (iv) deep learning methods: MLP and TabM \cite{gorishniy_tabm_2025}. We used a total of 112 small datasets, limited to 10 classes and up to 10000 samples, including binary and multiclass classification problems. The training data is split into $80\%$ training and $20\%$ calibration; hyperparameters are tuned on a held-out validation set, and final results are reported on the test set. For multiclass tasks, we report the weighted one-vs-one AUC. To stress-test model performance, we generate 20 synthetic datasets with high noise and weak class separation.

\section{Results}
\begin{figure}[H]
    \centering
    \includegraphics[width=0.9\linewidth]{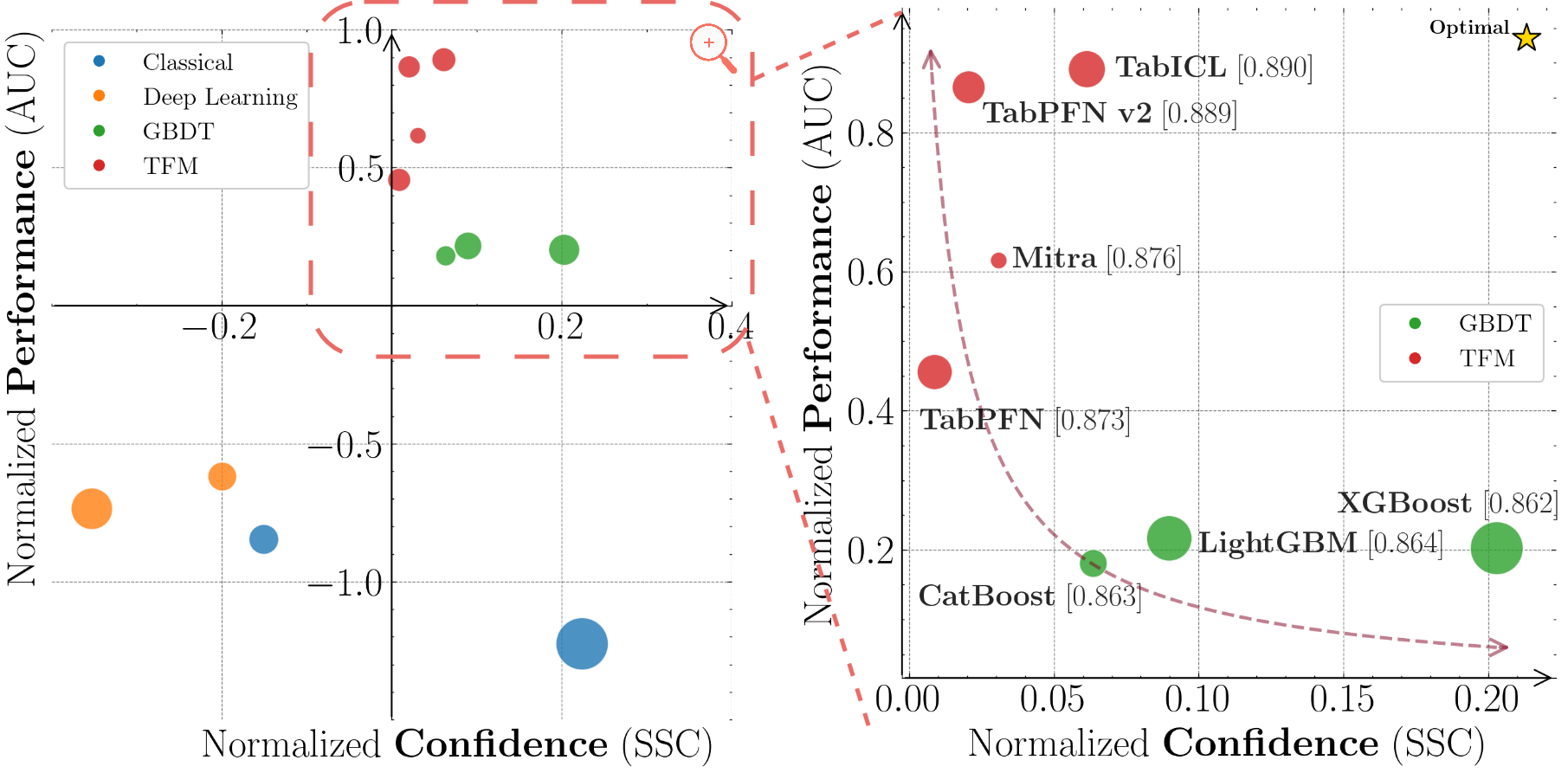}
    \caption{Performance--confidence trade-off (AUC vs.\ SSCS). Metrics are normalized per dataset and averaged; marker size denotes ECE. The dashed curve highlights the trade-off, and bracketed values report min--max normalized AUC. Non-normalized metrics are reported in Table~\ref{tab:metrics_by_model}.}
    \label{fig:tradeoff}
\end{figure}

Results are averaged over 15 seeds, with metrics reported in Table~\ref{tab:metrics_by_model} and Figure~\ref{fig:tradeoff}.

\begin{figure}[H]
\centering
\begin{minipage}[t]{0.5\textwidth}
 While Tabular Foundation Models (TFMs) achieve the highest AUC scores, their normalized Size-Stratified Coverage (SSC) remains significantly lower than that of GBDTs, revealing an imbalance between accuracy and reliability. This phenomenon is problematic as models with \textbf{poor conditional coverage tend to make overconfident errors}, producing narrow prediction sets that fail to include the true label. As highlighted by the blue versus red cell coloring, there is an \textbf{inverted trend between performance (AUC) and uncertainty quality (SSCS)}. Also, although logistic regression achieves good marginal calibration, its sharper prediction sets hurt conditional SSCS near non-linear boundaries.
\end{minipage}
\hfill
\begin{minipage}[t]{0.48\textwidth}
    \centering
    \vspace{0pt}
    \scalebox{0.75}{
    \begin{tabular}{lcc}
        \toprule
         Model & AUC~($\uparrow$) & SSCS~($\uparrow$) \\
        \midrule
        \multicolumn{3}{c}{Classical} \\
         knn & \cellcolor{blue!10}$0.815 \pm 0.022$ & \cellcolor{red!50}$0.637 \pm 0.064$ \\
         LogReg & \cellcolor{blue!14}$0.823 \pm 0.024$ & \cellcolor{red!33}$0.578 \pm 0.070$ \\
        \midrule
        \multicolumn{3}{c}{Deep Learning} \\
         tabm & \cellcolor{blue!19}$0.832 \pm 0.027$ & \cellcolor{red!31}$0.569 \pm 0.072$ \\
         mlp & \cellcolor{blue!20}$0.833 \pm 0.026$ & \cellcolor{red!14}$0.508 \pm 0.071$ \\
        \midrule
        \multicolumn{3}{c}{Foundation} \\
         PFN-v2 & \cellcolor{blue!49}$0.889 \pm 0.019$ & \cellcolor{red!10}$0.496 \pm 0.076$ \\
         tabicl & \cellcolor{blue!50}$0.890 \pm 0.019$ & \cellcolor{red!10}$0.494 \pm 0.076$ \\
         mitra & \cellcolor{blue!43}$0.877 \pm 0.021$ & \cellcolor{red!11}$0.500 \pm 0.075$ \\
         tabpfn & \cellcolor{blue!42}$0.874 \pm 0.021$ & \cellcolor{red!16}$0.517 \pm 0.074$ \\
        \midrule
        \multicolumn{3}{c}{GBDT} \\
         lightgbm & \cellcolor{blue!36}$0.864 \pm 0.022$ & \cellcolor{red!24}$0.544 \pm 0.072$ \\
         catboost & \cellcolor{blue!36}$0.864 \pm 0.022$ & \cellcolor{red!20}$0.532 \pm 0.075$ \\
         xgboost & \cellcolor{blue!35}$0.862 \pm 0.023$ & \cellcolor{red!23}$0.540 \pm 0.070$ \\
        \bottomrule
        \end{tabular}
    }
    \captionof{table}{Non-normalized AUC and SSCS for tabular models, averaged over 112 datasets and 15 seeds. The $\pm$ values denote standard deviation across datasets.}
    \label{tab:metrics_by_model}
\end{minipage}
\end{figure}

A closer analysis of dataset characteristics sheds light on when this trade-off becomes more pronounced. Under \textbf{extreme noise} or \textbf{low class separability}, TFMs produce high-variance, overconfident predictions. With non-Gaussian or skewed features, tree-based models remain robust, while TFMs lose calibration.

\begin{figure}[H]
\centering
\begin{minipage}[t]{0.4\textwidth}
    \centering
    \vspace{0pt}
    \scalebox{0.75}{
    \begin{tabular}{l c c}
    \toprule
    Group & AUC~($\uparrow$) & SSCS~($\uparrow$) \\
    \midrule
    Foundation & $\mathbf{0.924} \pm 0.005$ & $0.614 \pm 0.081$ \\
    GBDT & $0.889 \pm 0.007$ & $\mathbf{0.840} \pm 0.020$ \\
    \bottomrule
    \end{tabular}}
    \captionof{table}{Synthetic datasets}
    \label{tab:synth}
\end{minipage}
\hfill
\begin{minipage}[t]{0.5\textwidth}
\vspace{0pt}
Table \ref{tab:synth} present the results on the 20 synthetic datasets, with high noise and low class separability, averaged by different runs. As so, the performance-confidence contrast emerges clearly.
\end{minipage}
\end{figure}

Foundation models achieve the highest AUC, but their SSC remains comparatively low, indicating that they produce confident yet poorly calibrated predictions. These controlled results provide empirical evidence for a pattern that was also reinforced in the TALENT benchmark: while TFMs excel in accuracy, GBDTs consistently offer stronger and more trustworthy uncertainty estimates.

\section{Conclusion}

Our study exposes a fundamental trade-off between predictive accuracy and uncertainty reliability in tabular learning. While Tabular Foundation Models (TFMs) consistently outperform traditional models in terms of AUC, they exhibit weaker conditional coverage, indicating that their confidence does not faithfully reflect true predictive uncertainty. GBDTs, by contrast, maintain more stable uncertainty estimates, even at the cost of slightly lower accuracy.

Overall, \textbf{TFMs are best suited for well-curated, low-noise datasets} or few-shot scenarios where synthetic pretraining and in-context adaptation can leverage prior structure for higher accuracy. On small-to-medium datasets, GBDTs remain the most reliable choice for high-noise, heterogeneous, or complex feature distributions, offering better calibration and robustness.

Our analysis is limited to small and medium-scale datasets and to a model-agnostic conformal framework. Future work should evaluate TFMs under distribution shifts, explore model-specific uncertainty mechanisms (e.g., ensembling or Bayesian heads), and investigate how alternative pretraining distributions could improve calibration and reduce overconfidence.

\begin{footnotesize}
\bibliographystyle{unsrt}
\bibliography{references}
\end{footnotesize}

\end{document}